\documentclass{article}
\usepackage{ismir,amsmath,cite,url}
\pdfoutput=1
\usepackage{graphicx}
\usepackage{color}
\usepackage{float}

\usepackage{footnote}
\makesavenoteenv{tabular}
\makesavenoteenv{table}

\usepackage{titlesec}
\usepackage[toc,page]{appendix}

\title{Optical Music Recognition with Convolutional Sequence-to-Sequence Models}
\twoauthors
  {Eelco van der Wel} {University of Amsterdam \\ {\tt eelcovdw@gmail.com}}
  {Karen Ullrich} {University of Amsterdam \\ {\tt karen.ullrich@uva.nl}}
\begin{document}
\maketitle

\begin{abstract}
Optical Music Recognition (OMR) is an important technology within Music Information Retrieval. Deep learning models show promising results on OMR tasks, but symbol-level annotated data sets of sufficient size to train such models are not available and difficult to develop. We present a deep learning architecture called a Convolutional Sequence-to-Sequence model to both move towards an end-to-end trainable OMR pipeline, and apply a learning process that trains on full sentences of sheet music instead of individually labeled symbols. The model is trained and evaluated on a human generated data set, with various image augmentations based on real-world scenarios. This data set is the first publicly available set in OMR research with sufficient size to train and evaluate deep learning models. With the introduced augmentations a pitch recognition accuracy of 81\% and a duration accuracy of 94\% is achieved, resulting in a note level accuracy of 80\%. Finally, the model is compared to commercially available methods, showing a large improvements over these applications.
\end{abstract}
\section{Introduction}\label{sec:introduction}
Optical Music Recognition (OMR) is an application of recognition algorithms to musical scores, to encode the musical content to some kind of digital format. In modern Music Information Retrieval (MIR), these applications are of great importance. The digitization of sheet music libraries is necessary first step in various data-driven methods of musical analysis, search engines, or other applications where digital formats are required.

OMR is an active area of research in MIR, and a classically hard problem. OMR systems need to deal with a large range of challenges such as low quality scans, ambiguous notation, long range dependencies, large variations in musical font, and handwritten notation. Multiple commercial applications are available, each with their own strengths and weaknesses \cite{byrd2006prospects}, but the accuracy of these products is often too low to use without human supervision. An implementation that deals with these challenges in a satisfactory way has yet to be developed.

A traditional OMR system typically consists of multiple parts: score pre-processing, staff line identification and removal, musical object location, object classification and score reconstruction \cite{rebelo2012optical}. Each of these individual parts has its own difficulties, resulting in OMR systems with low confidence. More recently, there has been a trend towards less segmented systems involving machine learning methods, such as OMR without staffline removal \cite{pugin2007map} or without symbol segmentation \cite{shi2016end}. However, a major difficulty of these algorithms is the need for large amounts of training data. Typically, scores need to be annotated on musical symbol level to train such machine learning pipelines, but large corpora of sufficiently diverse symbol-annotated scores are difficult and expensive to produce \cite{rebelo2012optical}. 

In this study, we propose a novel deep learning architecture to both move towards an end-to-end trainable OMR pipeline, and greatly reduce the data requirements for training. This is achieved by using two common deep learning architectures: \textit{Convolutional Neural Networks} (CNN) and \textit{Recurrent Neural Networks} (RNN). Convolutional architectures have been a popular choice of algorithm in various MIR related tasks, due to the ability to learn local structures in images, and combining them to useful features. In our method, we use a CNN to learn a feature representation of the input scores. Continuing, a \textit{Sequence-to-Sequence} model \cite{sutskever2014sequence, cho2014learning} is used, which is a stack of two RNN's used commonly in machine translation tasks. This model directly produces a digital representation of the score from the learned representation by the CNN. The combination of these two architectures is called a \textit{Convolutional Sequence-to-Sequence} model.
By using a Sequence-to-Sequence architecture, we cast the problem of OMR as a translation problem. Instead of training on individual segmented symbols without context, full lines of sheet music are translated simultaneously. This approach has two major advantages; Firstly, by training the algorithm on full lines of sheet music, there is no need for symbol level annotated training data. This means that in principle any corpus of sheet music with corresponding digital notation could be used for training, opening up many new possibilities for data-driven OMR systems. Secondly, because in the proposed model each of the classically segmented OMR steps is done by a single algorithm, the model can use a large amount of contextual information to solve ambiguity and long-range dependency problems.

To train the proposed model, a large corpus of monophonic sheet music is generated from a MusicXML dataset as described in Section \ref{sec:data}. Additionally, in Section \ref{sec:augmentation} various types of image augmentations based on real-world scenarios are proposed to enhance the models flexibility to different kinds of fonts and varying score quality. Finally in Section \ref{sec:results}, the results of the method are discussed on both clean and augmented data, and the weaknesses of the model are examined.

\section{Related Work}
A starting point for any OMR research is the overview paper by Rebelo et al. \cite{rebelo2012optical}, which contains a complete introduction to OMR systems and a description of the current state of the field. The paper describes four main stages that are necessary for any OMR pipeline: Image pre-processing, musical symbol recognition, musical information reconstruction and construction of musical notation. The second component, as the name suggests, is where the main recognition work is done. Detecting and removing staff lines, segmenting individual symbols, and classifying symbols. Systems where steps are conducted by different methods we call segmented systems.

Not all methods follow this model, recent data-driven approaches suggest merging or omitting some of these segmented steps. An example of this is an approach suggested by Pugin et al. \cite{pugin2006optical, pugin2007map}, which applies \textit{Hidden Markov Models} (HMM) to the recognition stage, without performing staff line removal. Shi et al. \cite{shi2016end} incorporate a deep learning approach with Connectionist Temporal Classification function \cite{graves2006connectionist} as decoding mechanism. They pose a similar idea to the method proposed in this research, with a difference in the encoder mechanism. Instead of using both a CNN and RNN as encoder, only a CNN is used. This is less computationally expensive, but the additional RNN in the Sequence-to-sequence model can make the method proposed in this research more context aware.\\
Symbol classification involving neural networks has been researched by several authors \cite{rebelo2010optical, wen2015new}. Convolutional architectures have been used for different OMR sub-tasks, such as staff-line detection \cite{calvo2017staff} or symbol recognition \cite{pinheiro2016deep}.

In a different paper, Rebelo et al. \cite{rebelo2010optical} research the use of Deformable Templates \cite{jain1997representation} with various classifiers to make symbol recognition invariant to changes in musical font. This method is very similar to the Elastic Transformations \cite{simard2003best} used in this research. However, we decide to use Elastic Transformations for ease of use and application speed.

\section{Dataset}\label{sec:data}
The dataset used in this research is compiled from monophonic MusicXML scores from the MuseScore sheet music archive \cite{musescore}. The archive is made up of user-generated scores, and is very diverse in both content and purpose. As a result, the dataset contains a large variation in type of music, key signature, time signature, clef, and notation style.\\
To generate the dataset, each score is checked for monophonicity, and dynamics, expressions, chord symbols, and textual elements are removed. This process produces a dataset of about 17 thousand MusicXML scores. For training and evaluation, these scores are split into three different subsets. 60\% is used for training, 15\% for validation and 25\% for the evaluation of the models. A specification to reproduce the data set is publicly available online. 
\footnote{\url{https://github.com/eelcovdw/mono-musicxml-dataset}} \\

\subsection{Preprocessing}\label{sec:preprocessing}
From the corpus of monophonic MusicXML files, a dataset of images of score fragments and corresponding note annotations is created. Each MusicXML score is split into fragments of up to four bars, with a two bar overlap between adjacent fragments. The fragments are converted to sheet music using MuseScore \cite{musescore}, each image containing a single staff line. The corresponding labels are represented with a pitch and duration vector, containing all information about the notes and rests within the same four bars. Each musical symbol is represented with two values: a pitch, and a duration. Pitch values are specified by a MIDI pitch, and durations by quarterlength. In case of a rest, the pitch is a special rest indicator, which we indicate with $r$. The possible duration classes contain only the durations that can be specified by a single notehead. Notes with durations that require multiple noteheads are split into multiple notes. The first note will contain the pitch, and pitches of subsequent tied notes are replaced with a tie indicator, which we indicate with $t$

As an example, a quarter rest followed by a note with MIDI pitch 60 and a complex duration of a tied quarter note and a sixteenth note is notated as $((r, 1), (60, 1), (t, 0.25))$. Applying this method to the full score fragments produces the pitch and duration vector, and is a suitable representation for the model. A maximum of 48 events per fragment is used to put a limit on the sequence length the model has to decode. Finally, at the end of each pitch and duration vector an extra event is added to indicate the sequence has ended. This indicator is implemented as a rest with duration of zero quarter notes.

Each generated image is padded to the same width and height, and images containing notes with more than five ledger lines are discarded. These notes are extreme outliers and do not occur in normal notation. The resulting fragments have a dimension of $2261 \times 400$ pixels.

\begin{figure}[h]
 \centerline{
 \includegraphics[width=\columnwidth]{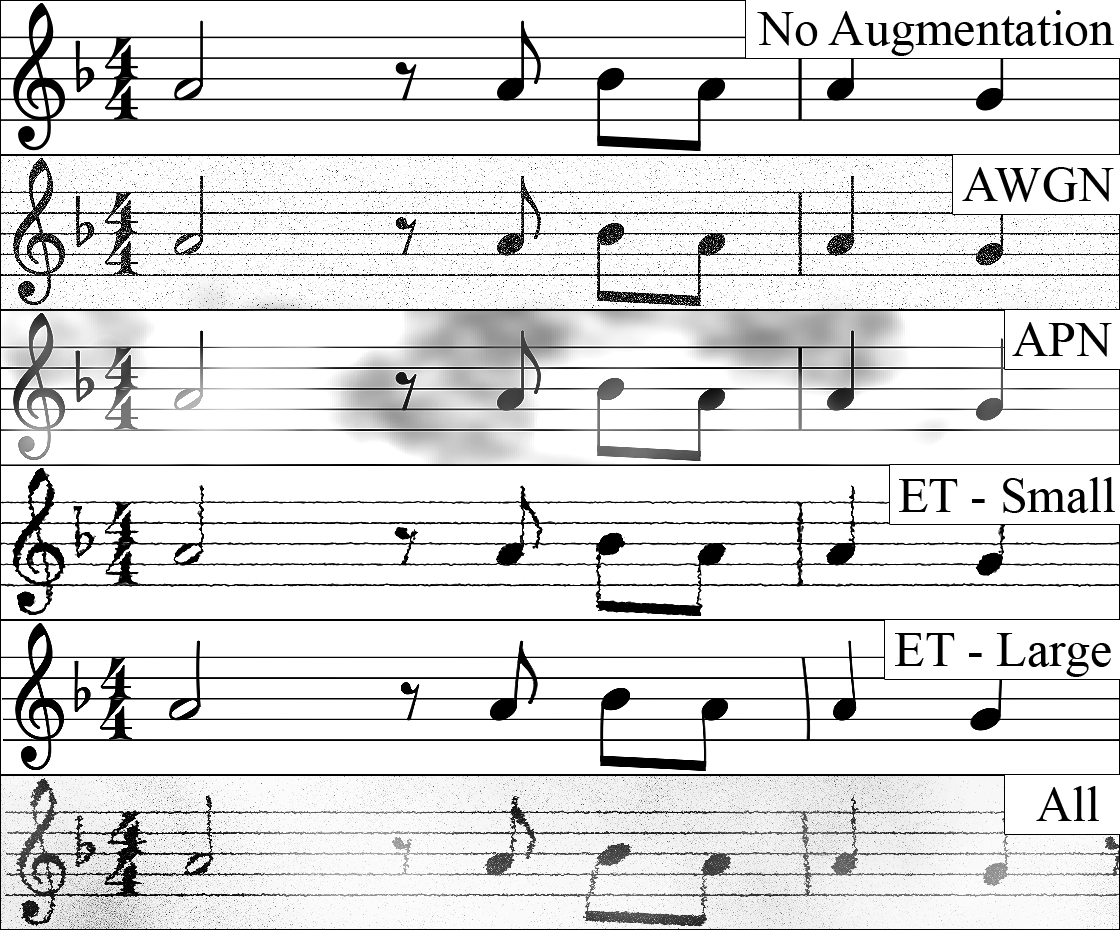}}
 \caption{An example of each of the used image augmentations from the evaluation dataset. From top to bottom: No Augmentations, Additive White Gaussian Noise (AWGN), Additive Perlin Noise (APN), Small scale Elastic Transformations (ET small), Large scale Elastic Transformations (ET large), and all combined augmentations.}
 \label{fig:aug}
\end{figure}
\subsection{Image Augmentation}\label{sec:augmentation}
The computer generated score fragments contain no noise or variation in musical symbols. To make the proposed model robust to lower quality inputs and different kinds of musical fonts we propose four different augmentations, each simulating a real world source of input noise. Additionally, for each augmentation, we choose two separate settings. For the augmented training data, the parameters are chosen such that the input sheet music is greatly deformed but still readable. For the augmented evaluation set, parameters are chosen such that they resemble real-world sheet music, with less deformation than the training data. The larger amount of deformation in training will force our model to learn to recognize musical symbol in any situation, and should improve the accuracy of our model on both non-augmented and augmented evaluation data.

A popular choice of augmentation is \textit{Additive White Gaussian Noise} (AWGN). This augmentation introduces a normally distributed random deviation in pixel intensities, to mimic noise introduced by low quality scans or photos. This noise has a mean $\mu$, which is chosen to be the same as the mean pixel intensity of the full dataset. The standard deviation $\sigma$ is different between our training and evaluation set. In the training set, the $\sigma$ of pixel intensities in our non-augmented data set is used. The evaluation set has a $\sigma$ of half that value.

The second type of noise augmentation used is \textit{Additive Perlin Noise} \cite{perlin2002improving}. Perlin noise is a procedurally generated gradient noise, that generates lighter and darker areas in the image at larger scales than AWGN. This effect mimics quality differences in parts of the score. Some symbols might be faded and parts of staff lines less visible, and dark areas in the image are created. The mean size of generated clouds is controlled by a frequency parameter. For each augmented score, this frequency is chosen to be a random value between the size of one note head and the mean width of a full bar, to generate noise structures at different scales. The maximum intensity of the noise in our training set is chosen to be a strength of 0.8. The evaluation set uses a maximum intensity of half this value.

The final two augmentations are achieved with \textit{Elastic Transformations} (ET) \cite{simard2003best}, which apply a smoothed field of local random affine transformations, resulting in wave-like displacements in the augmented image. An advantage of using this augmentation is that it applies a large range of possible affine and geometric transformations to each image, such as rotation, skewing, squeezing and stretching. This both enhances the diversity of the augmented data and alleviates the need to use manually defined geometric transformations.

Two parameters are used to control an elastic transformation: a strength factor $\sigma$, which reduces the strength of the distortion if a larger value is used, and a smoothing factor $\alpha$, which controls the scale of deformations. A very large $\alpha$ will apply a nearly linear translation to the image, while an $\alpha$ of zero applies fully random displacements on individual pixels.

The first type of Elastic Transformation is applied on very small scales, to change the characteristics of lines and smaller symbols. Lines might appear to be drawn by pencil or pen, and the edges of symbols become less defined. $\alpha$ is chosen to be a random value between 2 and 8, with a $\sigma$ of 0.5 for the training data, and a $\sigma$ of 2 for the evaluation data.

The second type of Elastic Transformation is applied on a large scale to change the shape and orientation of musical symbols. Barlines and note stems get skewed or bent, note heads can be compressed or elongated, and many new shapes are introduced in the score. This transformation mimics the use of different musical fonts, or even handwritten notation. An $\alpha$ between 2000 and 3000 is used, with a $\sigma$ of 40 for the training data, and 80 for the evaluation data. To maintain straight and continuous stafflines, the original algorithm is slightly adapted to reduce vertical translations of pixels by reducing the vertical component of transformations by 95\%.

In Figure \ref{fig:aug}, an example of each of these four augmentation is shown, with the setting used for generating the evaluation data. The last example shows a combination of all four augmentations.

\begin{figure*}[h!]
 \center
 \framebox{
 \includegraphics[width=16cm]{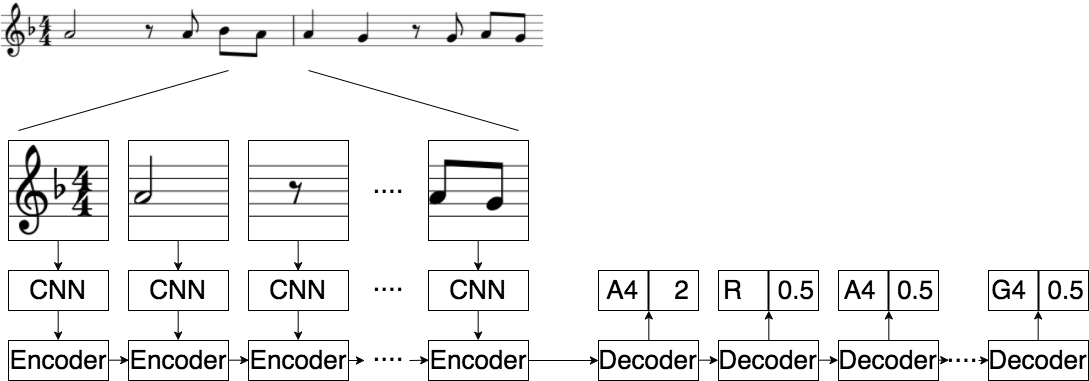}}
 \caption{A diagram of the proposed Convolutional Sequence-to-Sequence model. On the left, a score fragment is processed by a CNN and Encoder RNN to a fixed size representation. This representation is used by the decoder RNN to create a sequence of (pitch, duration) pairs.}
 \label{fig:model}
\end{figure*}
\section{Method}
We introduce the Convolutional Sequence-to-Sequence model as applied to OMR tasks, translating lines of sheetmusic to a sequence of (pitch, duration) pairs. Continuing, the training and evaluation methods are defined.
\subsection{Model}

We define a Convolutional Sequence-to-Sequence network as a stack of three components. First, a CNN encodes the input image windows to a sequence of vector representations. Then, an encoder RNN encodes the vector sequence to a fixed size representation, containing all information from the input score. Finally, a decoder RNN decodes the fixed size representation to a sequence of output labels. The following section describes each component in detail. Note that, while each component is described separately, the model will be trained as a single algorithm.

\textbf{Sliding window input}.
The image input of the algorithm is defined as a sequence of image patches, generated by applying a \textit{sliding window} over the original input score. The implementation has two separate parameters: the window width $w$ and window stride $s$. By varying $w$, the amount of information per window can be increased or decreased. $s$ defines how much redundancy exists between adjacent windows. Increasing the value of $w$ or decreasing the value of $s$ provides the model with more information about the score, but will raise the computational complexity of the algorithm. Thus when determining the optimal parameters, a balance has to be struck between complexity and input coverage. As a rule of thumb, we use a $w$ that is approximately twice the width of a notehead, and an $s$ of half the value of $w$. This will ensure that each musical object is shown in full at least once in an input window. This gives a $w$ of 64 pixels with a $s$ of 32.

\textbf{Convolutional neural network}.
To extract relevant features from the image patches, each patch is fed into a CNN. In this research, we keep the architecture of the CNN the same between different experiments, to ensure a fair comparison. First a \textit{max-pooling} operation of $3\times 3$ is applied on the input window for dimensionality reduction. Then, a convolutional layer of 32 $5\times 5$ kernels is applied, followed by a \textit{relu} activation and $2\times 2$ max-pooling operation. These three layers are repeated, and a fully-connected layer of 256 units with relu activation is applied, so each input for the encoder will be a vector of size 256.

\textbf{Sequence-to-Sequence network.}
After extracting a vector description of each image patch, the sequence of vectors is fed into a Sequence-to-Sequence network \cite{sutskever2014sequence, cho2014learning}. This architecture consists of two RNN's. The first RNN, the \textit{encoder}, encodes the full input sequence to a \textit{fixed size representation}. The second RNN, the \textit{decoder},  produces a sequence of outputs from the encoded representation. In the case of the OMR task, this sequence of outputs is the sequence of pitches and durations generated from the MusicXML files. For both encoder and decoder, a single \textit{Long Short-Term Memory} (LSTM) \cite{hochreiter1997long} layer is used with 256 units. To predict both the pitch and duration, the output of the decoder is split into two separate output layers with a \textit{softmax} activation and \textit{categorical cross-entropy} loss. 

A diagram of the full model is shown in \figref{fig:model}, where four input patches and output predictions are shown. On the left side, A sliding window is applied to a 2 bar score fragment. Each image patch is sequentially fed into the same CNN. This CNN is connected to the encoder network, creating a fixed size representation of the two input bars. The decoder uses this representation to produce the output sequence of (pitch, duration) pairs. Note that the second predicted pitch is an $r$ pitch, representing a rest symbol.

Using the described configuration, the model has an input sequence length of 70 windows and an output sequence length of 48 units. Shorter output sequences are padded to this maximum length and the loss function is masked after last element of the sequence. The number of pitch categories is 108, and the number of duration categories is 48. In total, the model contains approximately 1.67 million parameters.

\subsection{Training}
Six separate models are trained, one on each of the proposed augmented data sets: No augmentations, AWGN, APN, Small ET, large ET and all augmentations. Augmentations are applied during training, and will be different each time the network is presented with a training sample. All models are trained with a batch-size of 64 using the ADAM optimizer\cite{kingma2014adam}, with an initial learning rate of $8*10^{-4}$ and a constant learning rate decay tuned so the rate is halved every ten epochs. Each model is trained to convergence, taking about 25 epochs on the non-augmented dataset. A single Nvidia Titan X Maxwell is used for training, which trains a model in approximately 30 hours.
\newpage
\subsection{Evaluation Metrics}
On the evaluation data, three different metrics are calculated, similar to \cite{byrd2006prospects}:
\begin{itemize}
  \setlength{\itemsep}{1pt}
  \setlength{\parskip}{0pt}
  \setlength{\parsep}{0pt}
    \item Pitch accuracy, the proportion of correctly predicted pitches.
    \item Duration accuracy, the proportion of correctly predicted note durations.
    \item Note accuracy, the proportion of predicted events where both pitch and duration are correctly predicted.
\end{itemize}

The accuracy is measured over all notes before the stop indicator, and the stop indicator is not included in the calculation of accuracy. The model is not given any a priori knowledge about how many notes are in the input fragment, so a wrong number of notes could be predicted. In case of a shorter predicted sequence, the missing notes are automatically marked as incorrect. If the predicted sequence is longer than the ground truth, the additional predicted notes are cut and only the notes within the length of the ground truth are used. This method of measuring accuracy is quite strict, as an insertion or omission of a note in the middle of a sequence could mean subsequent notes are all marked as incorrect.  This should be kept in mind when evaluating the results of the model, and perhaps more descriptive metrics could be applied in future work.

\section{Results}\label{sec:results}

\subsection{Model Evaluation}
The six trained models are evaluated on both a clean evaluation set, shown in Table \ref{table:clean}, and augmented sets, shown in Table \ref{table:aug}. The augmented evaluation sets are generated by applying the augmentations the model was trained on to the full clean evaluation set, with the parameters described in Section \ref{sec:augmentation}.

The model trained on data with all augmentations is compared against two commercially available methods in Table \ref{table:com}, similar to Shi et al. \cite{shi2016end}. The comparison between the different methods on the clean dataset gives a baseline performance on digital scores, while the comparison on augmented data gives an indication of the difference of performance on real-world sheet music.
\begin{table}[h]
\begin{tabular}{|l|l|l|l|}
\hline
\textbf{\begin{tabular}[c]{@{}l@{}}Training \\ Augmentation\end{tabular}} & \textbf{\begin{tabular}[c]{@{}l@{}}Pitch \\ Accuracy\end{tabular}} & \textbf{\begin{tabular}[c]{@{}l@{}}Duration \\ Accuracy\end{tabular}} & \textbf{\begin{tabular}[c]{@{}l@{}}Note \\ Accuracy\end{tabular}} \\ \hline
None & 0.79 & 0.92 & 0.76 \\ \hline
AWGN & 0.79 & 0.92 & 0.77 \\ \hline
APN & \textbf{0.82} & 0.91 & 0.79 \\ \hline
ET - Small & 0.78 & 0.91 & 0.76 \\ \hline
ET - Large & 0.79 & \textbf{0.94} & 0.78  \\ \hline
All augmentations& 0.81 & \textbf{0.94} & \textbf{0.80} \\ \hline
\end{tabular}
\caption{Measured accuracy on non-augmented scores. The accuracy scores for augmentations with the highest positive impact are in bold.}
\label{table:clean}
\end{table}

\begin{table}
\begin{tabular}{|l|l|l|l|}
\hline
\textbf{\begin{tabular}[c]{@{}l@{}}Training \\ Augmentation\end{tabular}} & \textbf{\begin{tabular}[c]{@{}l@{}}Pitch \\ Accuracy\end{tabular}} & \textbf{\begin{tabular}[c]{@{}l@{}}Duration \\ Accuracy\end{tabular}} & \textbf{\begin{tabular}[c]{@{}l@{}}Note \\ Accuracy\end{tabular}} \\ \hline
AWGN & 0.79 & 0.90 & 0.75 \\ \hline
APN & 0.81 & 0.89 & 0.76 \\ \hline
ET - Small & 0.78 & 0.89 & 0.74 \\ \hline
ET - Large & 0.78 & 0.94 & 0.75  \\ \hline
All augmentations& 0.79 & 0.92 & 0.77 \\ \hline
\end{tabular}
\caption{Measured accuracies on scores with augmentations. Each model trained on different augmented data is evaluated on an evaluation set with corresponding augmentations.}
\label{table:aug}
\end{table}

\begin{table}
\begin{tabular}{|l|l|l|l|}
\hline
\textbf{Model} & \textbf{Clean} & \textbf{Augmented} \\ \hline
Capella Scan 8 \footnote{\url{https://www.capella-software.com/}}& 0.53 & 0.14 \\ \hline
Photoscore 8 \footnote{\url{http://www.neuratron.com/}} & 0.61 & 0.09 \\ \hline
CS2S & 0.80 & 0.77 \\ \hline
\end{tabular}
\caption{A comparison of accuracy between the proposed model (CS2S) and two popular commercially available methods.}
\label{table:com}
\end{table}

\subsection{Evaluation of Model Difficulties}
To examine the difficulties the model has on different kinds of scores, three additional evaluations are performed on different subsets of the evaluation data.
\begin{figure}[h]
 \center
 \includegraphics[width=0.9\columnwidth]{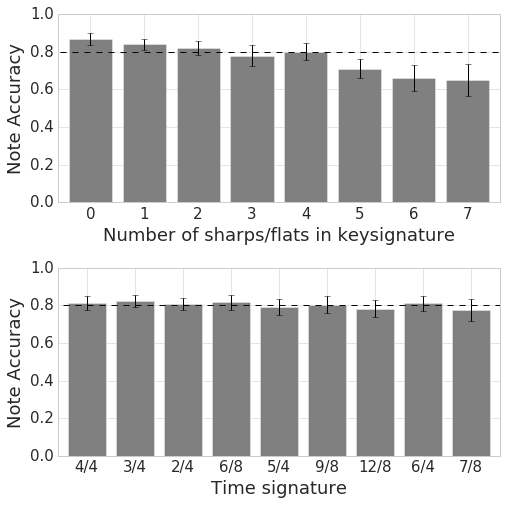}
 \caption{Top: The note-level accuracy for each number sharps/flats in the key signature. Bottom: The note-level accuracy for the most common time signatures. The dotted lines indicate the mean accuracy of 0.80.}
 \label{fig:sig}
\end{figure}

First, an investigation into the impact of key signature on the note level accuracy on the non-augmented evaluation data is conducted. Just like human performance, the added complexity of many sharps or flats in the key signature of a fragment impacts the accuracy of the model. The results of this experiment are displayed in Figure \ref{fig:sig} (top). At zero sharps or flats, the reported accuracy is 0.86, achieving 0.06 higher than the mean accuracy of 0.80. With more than 4 sharps or flats in the key signature the note accuracy starts diminishing, down to a minimum of 0.66 for key signatures with seven sharps or flats.

Continuing, the nine most common time signatures and their accuracies are examined. While the output notation does not encode any direct information about time signature, the model could use structural information imposed by the time signature on the score to aid in note recognition. This evaluation will both look at if that is the case, and investigate which time signatures are potentially more difficult to transcribe. The results in Figure \ref{fig:sig} (bottom) do not show a significant difference between the measured accuracy of different time signatures. The complex time signatures of $7/8$ and $5/4$ both are slightly less accurate, but this observation could be caused by a random deviation, or by features correlating with complex time signatures such as number of notes in a fragment.

\begin{figure}
 \center
 \includegraphics[width=\linewidth]{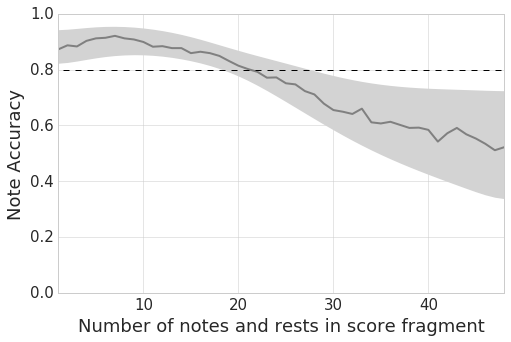}
 \caption{The mean note-level accuracy for each number of notes per fragment, with a confidence interval at a 95\% level fit with a Gaussian Process.}
 \label{fig:n_notes}
\end{figure}

As a final evaluation, we look at the correlation between number of notes in a fragment and accuracy. The model capacity and the representation between encoder and decoder are of a fixed size, which forces the model to represent more notes in the same space for fragments with a higher note density. This higher density could cause a loss in accuracy. Figure \ref{fig:n_notes} shows clear evidence that this is the case; fragments containing more than 25 notes have a significantly lower accuracy than the measured mean.

\section{Discussion}
We propose the Convolutional Sequence-to-Sequence model to deal with the difficulties OMR presents for learning systems. By using an end-to-end trainable sequential model, we completely move away from segmented symbol recognition, and perform the full OMR pipeline with a single algorithm.
By incorporating Sequence-to-Sequence models into OMR, there are many new possibilities for obtaining development data. We view this aspect as the largest advantage the proposed method has over segmented models, as the acquisition of quality training data can be a limiting factor. The proposed model shows that it is robust to noisy input, an important quality for any OMR model. Additionally, the experiments show that it can deal with the large scale Elastic Transformations that essentially change the musical font. In future research, this aspect could be expanded to include handwritten notation.

A weakness of the model is pitch classification. Pooling operations introduce a degree of translation invariance, we hypothesize this invariance reduces the pitch recognition accuracy by discarding information about symbol position. However, omitting pooling operations from the model would greatly reduce the dimensionality reduction performed by the CNN.
We propose incorporating a combination of convolutional layers and fully connected layers as a possible solution.

Furthermore, on more complex scores the model performs significantly worse. Both the number of sharps or flats in the key signature and the note density in the score fragment play a large role in the prediction accuracy. In future work, these problems could be addressed in multiple ways. A separate key signature recognition could be performed, and given as additional information to the model. This would take away some of the long range computations the key signature introduces and could improve the results on more complex scores. 

The difficulty of translating long sequences with Sequence-to-Sequence models is a well studied problem \cite{sutskever2014sequence, cho2014learning}. For longer sequences, the model needs to encode more information in the same fixed size representation, reducing the amount of storage available per note. A possible solution for this difficulty is proposed by Bahnadau et al. \cite{bahdanau2014neural}: they replace the fixed size representation between encoder and decoder with an attention mechanism, a method that essentially performs a search function between the two networks. This mechanism has shown improvements to Sequence-to-Sequence models in neural machine translation, and could be used in the proposed method to alleviate some of the problems introduced with long sequences and long range dependencies.

The experiments performed in this research are exclusively on monophonic scores. The current representation of (pitch, duration) pairs does not allow for polyphonic note sequences, and in order to apply the model to polyphonic OMR tasks this representation needs to be adapted. A possible representation could be produced by using a method close like the MIDI-standard or piano roll representation.

Finally, we propose that the Convolutional Sequence-to-Sequence model could be applied to tasks outside of OMR that translate a spatial sequential representation to a sequence of labels. Within MIR, tasks like Automatic Music Transcription can be considered as such a task, where a representation of an audio signal is converted to a sequence of pitches and durations. Outside of MIR, tasks like video tagging or Optical Character Recognition are similar examples.

\section{Acknowledgements}
Many thanks to the reviewers for their valuable feedback, and to MuseScore for providing access to the material available in their archive. This research has been partially funded by Google.



\bibliography{references.bib}

\begin{thebibliography}{10}

\bibitem{musescore}
Musescore.
\newblock \url{https://musescore.org/}.

\bibitem{bahdanau2014neural}
Dzmitry Bahdanau, Kyunghyun Cho, and Yoshua Bengio.
\newblock Neural machine translation by jointly learning to align and
  translate.
\newblock {\em arXiv preprint arXiv:1409.0473}, 2014.

\bibitem{byrd2006prospects}
Donald Byrd and Megan Schindele.
\newblock Prospects for improving omr with multiple recognizers.
\newblock In {\em ISMIR}, pages 41--46, 2006.

\bibitem{calvo2017staff}
Jorge Calvo-Zaragoza, Antonio Pertusa, and Jose Oncina.
\newblock Staff-line detection and removal using a convolutional neural
  network.
\newblock {\em Machine Vision and Applications}, pages 1--10, 2017.

\bibitem{cho2014learning}
Kyunghyun Cho, Bart Van~Merri{\"e}nboer, Caglar Gulcehre, Dzmitry Bahdanau,
  Fethi Bougares, Holger Schwenk, and Yoshua Bengio.
\newblock Learning phrase representations using rnn encoder-decoder for
  statistical machine translation.
\newblock {\em arXiv preprint arXiv:1406.1078}, 2014.

\bibitem{graves2006connectionist}
Alex Graves, Santiago Fern{\'a}ndez, Faustino Gomez, and J{\"u}rgen
  Schmidhuber.
\newblock Connectionist temporal classification: labelling unsegmented sequence
  data with recurrent neural networks.
\newblock In {\em Proceedings of the 23rd international conference on Machine
  learning}, pages 369--376. ACM, 2006.

\bibitem{hochreiter1997long}
Sepp Hochreiter and J{\"u}rgen Schmidhuber.
\newblock Long short-term memory.
\newblock {\em Neural computation}, 9(8):1735--1780, 1997.

\bibitem{jain1997representation}
Anil~K Jain and Douglas Zongker.
\newblock Representation and recognition of handwritten digits using deformable
  templates.
\newblock {\em IEEE Transactions on Pattern Analysis and Machine Intelligence},
  19(12):1386--1390, 1997.

\bibitem{kingma2014adam}
Diederik Kingma and Jimmy Ba.
\newblock Adam: A method for stochastic optimization.
\newblock {\em arXiv:1412.6980}, 2014.

\bibitem{perlin2002improving}
Ken Perlin.
\newblock Improving noise.
\newblock In {\em ACM Transactions on Graphics (TOG)}, volume~21, pages
  681--682. ACM, 2002.

\bibitem{pinheiro2016deep}
Roberto~M Pinheiro~Pereira, Caio~EF Matos, Geraldo Braz~Junior, Jo{\~a}o~DS
  de~Almeida, and Anselmo~C de~Paiva.
\newblock A deep approach for handwritten musical symbols recognition.
\newblock In {\em Proceedings of the 22nd Brazilian Symposium on Multimedia and
  the Web}, pages 191--194. ACM, 2016.

\bibitem{pugin2006optical}
Laurent Pugin.
\newblock Optical music recognitoin of early typographic prints using hidden
  markov models.
\newblock In {\em ISMIR}, pages 53--56, 2006.

\bibitem{pugin2007map}
Laurent Pugin, John~Ashley Burgoyne, and Ichiro Fujinaga.
\newblock Map adaptation to improve optical music recognition of early music
  documents using hidden markov models.
\newblock In {\em ISMIR}, pages 513--516, 2007.

\bibitem{rebelo2010optical}
Ana Rebelo, G~Capela, and Jaime~S Cardoso.
\newblock Optical recognition of music symbols.
\newblock {\em International journal on document analysis and recognition},
  13(1):19--31, 2010.

\bibitem{rebelo2012optical}
Ana Rebelo, Ichiro Fujinaga, Filipe Paszkiewicz, Andre~RS Marcal, Carlos
  Guedes, and Jaime~S Cardoso.
\newblock Optical music recognition: state-of-the-art and open issues.
\newblock {\em International Journal of Multimedia Information Retrieval},
  1(3):173--190, 2012.

\bibitem{shi2016end}
Baoguang Shi, Xiang Bai, and Cong Yao.
\newblock An end-to-end trainable neural network for image-based sequence
  recognition and its application to scene text recognition.
\newblock {\em IEEE Transactions on Pattern Analysis and Machine Intelligence},
  2016.

\bibitem{simard2003best}
Patrice~Y Simard, David Steinkraus, John~C Platt, et~al.
\newblock Best practices for convolutional neural networks applied to visual
  document analysis.
\newblock In {\em ICDAR}, volume~3, pages 958--962. Citeseer, 2003.

\bibitem{sutskever2014sequence}
Ilya Sutskever, Oriol Vinyals, and Quoc~V Le.
\newblock Sequence to sequence learning with neural networks.
\newblock In {\em Advances in neural information processing systems}, pages
  3104--3112, 2014.

\bibitem{wen2015new}
Cuihong Wen, Ana Rebelo, Jing Zhang, and Jaime Cardoso.
\newblock A new optical music recognition system based on combined neural
  network.
\newblock {\em Pattern Recognition Letters}, 58:1--7, 2015.

\end{thebibliography}

\end{document}